\documentclass{article}

\usepackage{arxiv}

\usepackage[utf8]{inputenc} 
\usepackage[T1]{fontenc}    
\usepackage{hyperref}       
\usepackage{url}            
\usepackage{booktabs}       
\usepackage{amsfonts}       
\usepackage{nicefrac}       
\usepackage{microtype}      
\usepackage{lipsum}		
\usepackage{graphicx}
\usepackage{natbib}
\usepackage{doi}

\graphicspath{{./figs/}}
\DeclareGraphicsExtensions{.pdf,.jpeg,.jpg,.png}
\usepackage{subcaption}
\usepackage{url}

\title{Leaf Segmentation and Counting with Deep Learning: on Model Certainty, Test-Time Augmentation, Trade-Offs}

\author{ {Douglas P. S. Gomes} \\
    School of Computing and Mathematics\\
	Charles Sturt University\\
	Melbourne, Australia \\
	\texttt{douglas.uf@gmail.com} \\
	\And
	{Lihong Zheng} \\
    School of Computing and Mathematics\\
	Charles Sturt University\\
	Wagga Wagga, Australia \\
	\texttt{lzheng@csu.edu.au} \\
}

\date{}

\renewcommand{\shorttitle}{Leaf Segmentation and Counting with Deep Learning}

\hypersetup{
pdftitle={Leaf Segmentation and Counting with Deep Learning},
pdfsubject={CV},
pdfauthor={Douglas P. S. Gomes},
pdfkeywords={Leaf segmentation, counting, deep learning, augmentation},
}

\begin{document}
\maketitle

\begin{abstract}
Plant phenotyping tasks such as leaf segmentation and counting are fundamental to the study of phenotypic traits. Since it is well-suited for these tasks, deep supervised learning has been prevalent in recent works proposing better performing models at segmenting and counting leaves. Despite good efforts from research groups, one of the main challenges for proposing better methods is still the limitation of labelled data availability. The main efforts of the field seem to be augmenting existing limited data sets, and some aspects of the modelling process have been under-discussed. This paper explores such topics and present experiments that led to the development of the best-performing method in the Leaf Segmentation Challenge and in another external data set of Komatsuna plants. The model has competitive performance while been arguably simpler than other recently proposed ones. The experiments also brought insights such as the fact that model cardinality and test-time augmentation may have strong applications in object segmentation of single class and high occlusion, and regarding the data distribution of recently proposed data sets for benchmarking. 
\end{abstract}


\section{Introduction}
\label{sec:intro}

As part of the broader field of plant phenotyping, the tasks of leaf segmentation and counting have been receiving increasing attention with some distinguishable work focusing solely on either or both of these tasks (\cite{kuznichov2019data, ward2020scalable, zhu2018data}). This is not necessarily surprising since the analysis of phonetic traits can play a role in the advancement of plant science and in the aspects of breeding and crop management. Nevertheless, there are potential underlying factors for such growth besides the importance of the subject. The potentially most relevant one has to do with possible benefits from applying computer vision-based techniques to the tasks of plant phenotyping, especially ones in deep learning, which also experienced tremendous growth in the past years. Such benefits are mainly composed of moving from manually or semi-automatically labelling, which can be much onerous, expensive and low-throughput, to fully end-to-end segmentation (leaf area measuring) and counting in a high-throughput manner.

These plant-phenotyping tasks relate to broader problems in computer vision: instance segmentation (leaf segmentation) and object detection (leaf counting). It is not surprising, therefore, that deep learning has been so prevalent in works addressing these tasks in plant phenotyping since it has also been responsible for new benchmarking in their respective computer-vision tasks. Leaf segmentation can be seen as a particular case of instance segmentation where the number of classes is equal to one. Leaf counting relates to object detection in the sense that the number of leaves on an image can be given by the number of instances detected from such a single-class object. It is important to note that localization (segmentation) and detection have an intrinsic trade-off, which is usually regulated by a detection threshold that influences how many instances are counted as positive detections and how well-localized they are.

As other maturing fields in this current period of higher accessibility to supervised-learning algorithms and increasing volume of works, there is a need for standardized benchmarks and data sets to objectively compare different solutions. To this issue, one can find remarkable efforts from some research groups that made their data publicly available (\cite{uchiyama2017easy,MinerviniPRL2015}). However, as will be further discussed here, all data sets carry relevant bias and one should make an effort to measure how the results from one data set will actually generalise to others. Moreover, as the use of deep learning is relatively recent in this field, there are some under-discussed topics like the parameters of the model that relates to certainty, which can have great implications and trade-offs in tasks like segmentation and instance detection. 

Two important data sets available in the literature are named here as the CVPPP (\cite{MinerviniPRL2015}) and the Komatsuna (\cite{uchiyama2017easy}). The CVPPP is the most popular and firstly introduced in the Computer Vision Problems in Plant Phenotyping (CVPPP) 2014 workshop. The authors not only released the well-annotated data set of Arabidopsis and Tobacco images taken from the top but also specific metrics for benchmarking solutions. Later on, the team also set up a CodaLab server \cite{codalab} as the Leaf Segmentation Challenge (LSC), which was a test set where the ground truth labels are not shared so solutions can be effectively compared in the released metrics. The second data set comprises images of Komatsuna also taken from the top with ground truth, firstly presented in the ICCV Workshop on Computer Vision Problems in Plant Phenotyping in 2017. This data set, as the validation server in the CodaLab, is used here strictly as a test set to obtain metrics of generalisation and indirectly measure the bias of the CVPPP data set (used for learning), which is now a benchmark of the field.

One interesting point to the work proposed here is that past works presenting competitive results on the CVPPP data set have been achieving it through ingenious data augmentation techniques (\cite{gomes_zheng_2020}). Most involve the generation of synthetic data via elaborate methods ranging from copy-and-paste methods (\cite{kuznichov2019data}) to generative artificial networks (\cite{zhu2018data}). Although some are simpler than others, they all represent expensive techniques which are not obviously worth the investment of time and resources. 
The results presented here attest to this fact by presenting models that rank first in the LSC without using synthetic data generation methods and accessible computer-vision frameworks. The results are achieved by simple data augmentation and evaluation techniques, representing an inexpensive alternative to other more intricate methods. As previously mentioned and important point, the effect of the detection threshold on the trade-off of localization and instance detection is also performed. They show that there are some disadvantages when adopting metrics such as dice score and leaf count for benchmarking. Moreover, experiments with models of different depth and cardinality are also performed to attest for their different performance gain and losses. They are part of the analysis of bias in the CVPPP data set and are illuminating as to their influence in overfitting and performance on the tasks of segmentation and instance counting. In short, the following contributions are proposed: 

\begin{itemize}
    \item Strategies for producing leaf segmentation and counting models presenting state-of-the-art results while simpler.
    \item An analysis of the effect of different detection thresholds and its trade-off in segmentation and counting.
    \item The effect of model size, cardinality, and test-time augmentation in leaf segmentation.
    \item Experiments and commentary on the popular CVPPP test data set regarding its bias and potential influence in model generalisation.
\end{itemize}

\section{Related work}
\label{sec:rel_work}

There are some noteworthy works that comprise the same scope as this paper and should therefore be commented. The field of plant phenotyping is much larger than such a scope, so related works are going to be the ones considering similar tasks to the ones considered here, i.e. leaf segmentation and counting. To make the results as comparable as possible, a focus will be given to works that use the same data set as adopted in this paper. That does not necessarily limit the number of works, as the CVPPP has become a benchmark to these tasks and even used in works proposing instance segmentation generic techniques (\cite{ren2017end}). However, this does limit the works mainly to the ones related to the CVPPP as only two other papers adopt the Komatsuna data set, as identified by the present authors. A noteworthy point regarding recent works presented in the past two to three years is that many share one particular characteristic: the main contribution and alleged boost in performance come from data augmentation applications. They range from simple techniques to intricate pipelines to generate synthetic data to augment training. The fact that they are so prevalent even inspired the authors of this paper to write a review describing these methods (\cite{gomes_zheng_2020}).

A potentially simple technique for data augmentation, for example, is creating synthetic data by selecting related backgrounds and pasting cut instances of plants, or seeds, in random orientations. The work by \cite{kuznichov2019data} proposes such a technique on the CVPPP data set as a structured collage, in which the pasting follows specific rules like the centering the leaves in a pot background to resemble a rosette image taken from the top.  Another work by \cite{toda2019learning} presented the same technique in a seed segmentation task, obtaining significant performance gains.

Other more intricate techniques can involve using Generative Artificial Network (GANs) or complicated modelling pipelines to generate synthetic data. As an example of using such an approach, \cite{ubbens2018use} used an L-systems-based simulator software to model plants with traits variable according to probabilities distributions. It allegedly presented evidence of the transferability in learning from plant modelling to the tasks of leaf counting. Moreover, the authors in \cite{ward2020scalable} presented a full-dedicated pipeline comprising background processing, texture generation and plant assembly. The authors claim to have surpassed state-of-the-art results at the time, and that their technique would help to bridge the gap between species. They were also one of the few that submitted their results to the LSC server and thus more objectively compared. Regarding the methods using GANs to augment training data (\cite{valerio2017arigan, zhu2018data}), they have been mostly used to address the task of leaf counting. They seem to be effective at boosting performance but suffer from being complex to implement and highly specific, focusing on a single plant species from which images are generated.

Given the number of papers using the approach of data augmentation to address the tasks of leaf segmentation and counting, one could argue that they are effective, but they might also be too domain-specific. When augmenting training data with GANs, for example, the images generated were from Arabidopsis, and no evidence was provided that it would help with any other species. The same is true for the works using modelling and collage augmentation. It is one of the reasons why an external data set was added to the validation step to the work proposed here. An exception of such critique is the work by \cite{ward2020scalable}, which had the goal of generalising to other species with their data generation pipeline; they tested their method with two data sets, one of which was the one used here. Since they were also one of the few that submitted their evaluation to the LSC server, it is one of the works that can be most objectively compared. This point is important because, as most of these methods are not officially present in the LSC challenge, the comparison of the results will not be strict and will be related to the results presented in their paper. Often such results are extracted from splits made in the CVPPP data set from A1-A4 into training and testing sets. Hence, the results presented here, which were extracted from a test set in the server with no ground truth available, should receive a high weight regarding validation and generalisation ability. One honourable mention is the recent method by \cite{wu2020improving}, which presents a whole new architecture based on the U-Net and clustering pixels with the same characteristics to obtain leaf masks. Nevertheless, the results presented here also overperform their method in the leaf segmentation task, like all other mentioned works.

\section{Data and methods}
\label{sec:data_methods}

This work presents experiments testing hypotheses around the tasks of leaf segmentation and counting and proposes techniques that result in a computer vision model achieving results comparable to the state of the art in such tasks. This section describes the two independent data sets, the experiments performed to assess the hypotheses, and the techniques that led to the competitive results discovered throughout the investigation on these tasks.

\subsection{Data sets}
The analysis proposed here is based primarily on the CVPPP data set while the validation of the presented hypotheses is performed on the Leaf Segmentation Challenge server and on an independent similar data set of a different plant species (Komatsuna). As previously noted, the importance of the CVPPP is given by its adoption in many works in the field of plant phenotyping, especially regarding the problems of leaf segmentation and counting. The challenge server then becomes an interesting validation tool as it comprises a validation set that no one has access to the ground-truth masks, although not many authors submit their results there for comparison to other works. Nevertheless, as some of the insights presented here are comments on the data distribution of the CVPPP and its effect on the tasks of segmentation and counting, an external data set is needed to test and validate the hypothesis.

\subsubsection{The CVPPP data set and challenge}
\begin{figure}[t]
  \centering
  \begin{subfigure}[t]{0.3\textwidth}
    \includegraphics[width=\textwidth]{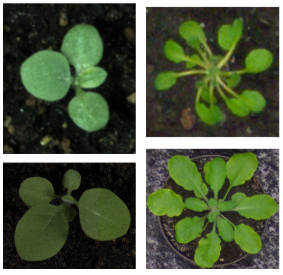}
    \caption{Examples of plants from the CVPPP dataset.}
  \end{subfigure}             
  \begin{subfigure}[t]{0.3\textwidth}
    \includegraphics[width=\textwidth]{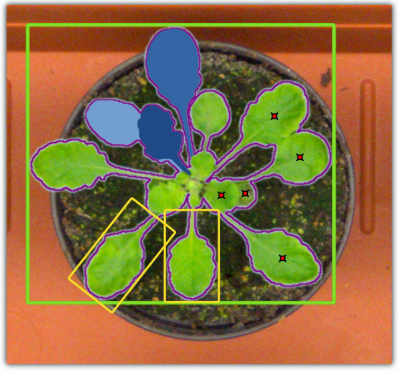}
    \caption{Examples of possible labels.}
  \end{subfigure}             
  \caption{Examples of plant images in the CVPPPP dataset.}
  \label{fig:cvppp}
\end{figure}

The CVPPP data set represents perhaps that largest effort to address the ongoing problem of lacking benchmark data sets and metrics for specific tasks such as leaf segmentation and counting on a controlled environment. First presented in 2014, but updated in 2017, the data set comprises images of mainly Arabidopsis and a small portion of Tobacco plants and can be accessed by \cite{PlantPhenotypingDatasets2015}. The authors stated that the images have been collected from several sites from growth chamber experiments, and it is divided into four groups named A1 to A4. There are 810 images with ground-truth masks available, from which 783 are from Arabidopsis and 27 from Tobacco, showing that the later represents only a small part of that data. These images are all taken from the top, probably as an attempt to mitigate occlusion, as illustrated in Fig. \ref{fig:cvppp}.

The organizers of this data set made the critical decision of creating a separate data set, A5, and uploading it to an evaluation server in CodaLAB labelled as the Leaf Segmentation Challenge (LSC). They also set up metrics for the tasks trained on the data set, from which two central are adopted in this paper: Symmetric Best Dice (SBD) for segmentation and Absolute Difference in Count (abs. DiC) for counting. These efforts paid off with the data set now being used as a benchmark in many recent papers (\cite{ren2017end,kuznichov2019data,ward2020scalable,zhu2018data}). Nevertheless, although many authors have been using it as a training and test set, a few have submitted to the challenge server. The model presented in this paper, which has used all A1-A4 images in training and tested in the A5 (server), currently ranks first in the challenge leaderboard.

\subsubsection{Komatsuna data set}

\begin{figure}[t]
  \centering
  \begin{subfigure}[t]{0.3\textwidth}
    \includegraphics[width=\textwidth]{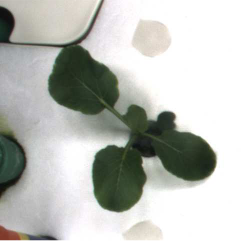}
    \caption{}
  \end{subfigure}             
  \begin{subfigure}[t]{0.3\textwidth}
    \includegraphics[width=\textwidth]{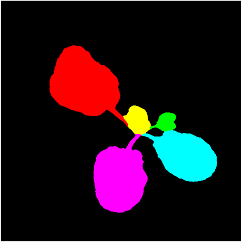}
    \caption{}
  \end{subfigure}             
  \caption{Examples of a plant image in the Komatsuna data set (a) and its ground-truth masks (b).}
  \label{fig:komatsuna}
\end{figure}

In a similar manner to the CVPPP data set, the Komatsuna (\cite{uchiyama2017easy}) comprises images of plants taken from the top but now on a different species, which is hoped to help attest generalization. Although the data set was conceptualized to help address 3D phenotyping problems, it is mainly used here as a test set only due to its similarities to the CVPPP (training set), as illustrated in Fig. \ref{fig:komatsuna}. Such a test set is composed of 271 images, and the evaluation is performed in the same metrics as used in the CVPPP testing. One final feature of this data set is that the images contain plants in many different stages of growth, as the data set was also conceptualized with this goal in mind.

\subsection{Detection threshold effect}

The detection threshold is a fundamental part of a classification or decision-making system and can have severe effects on its output. This effect is more explicitly illustrated by the Receiver Operating Characteristic (ROC) curve, which plots the false versus true positive rate. It shows that there is a trade-off between the number of detection a system makes and how many are actually true. For example, one could trivially have a maximum true positive rate by just classifying all examples as positive, but that would also lead to a very large false-positive rate. In computer vision, this concept is often discussed in terms of precision and recall. The effect that it has on object detection is remarkable: if one sets a low detection threshold, many candidates of objects will be selected, but a larger number will be miss detections or have worse localization (e.g.  bounding boxes). This is mainly a problem of certainty in the decision-making system and previously referred to in the literature as cost-sensitive analysis (\cite{elkan2001foundations}). 

Previous works proposing algorithms for segmenting leaves on the CVPPP have been using mainly deep learning, but also a specific method for computer vision called Mask R-CNN (\cite{he2017mask}). Such a method contrasts with other object detection algorithms for using a two-step process rather than being one-shot like methods such as YOLO (\cite{redmon2016you}). 
For the particular case on the Mask R-CNN framework \cite{he2017mask}, the trade-off mentioned in the last paragraph can come into play in the mask proposal head, which is the addition to the classic Faster R-CNN architecture. 
This head produces a fixed-sized mask for every region of interested created in the first stage, and the detections are attested when the intersection over union (IoU) value is higher than an arbitrary threshold.
Therefore, such an effect is relevant to the work presented here since recent methods for leaf segmentation often does not discuss the effect that different thresholds can have on the final output. To address such a gap, experiments with different thresholds were performed here and analysed in the Mask R-CNN framework. The results reveal a significant impact in the dice score when segmenting leaves, attesting to the hypothesis that the discussion about the detection threshold should not be neglected in works addressing this task. Simple changes boosted the performance of simple methods to state-of-the-art results but at the expensive trade-off with the leaf counting metric. Moreover, these experiments also bring relevant insights to the data sets used when results from the CVPPP and Komatsuna are compared, especially regarding their distribution and potential for overfitting.

\subsection{Model depth and cardinality}

Although it is somewhat expected that deeper models often lead to better performance, it is important to test such hypotheses, especially when the size of these data sets are put in context. The CVPPP data set, used here for training, comprises less than a thousand images, hence arguably prone to overfitting and the reason why so many previous works have been proposing data augmentation techniques with it. To test the effect of model depth, two popular variations of the Mask R-CNN are tested here, one having a 50-layer backbone while the other having 101 layers. 

To make the experiments potentially more insightful, not only two different depths were compared but also a variation with different cardinality on the 101-layer backbone was added. This addition resulted in two models having a ResNet backbone (\cite{he2016deep}), one with 50 and the other with 101 layers, and one model having a ResXNet backbone (\cite{xie2017aggregated}). The difference between these two types of backbone is in the size of transformations, which the authors call cardinality (\cite{xie2017aggregated}), but they do have the same dimensions regarding depth and width. The results from such comparison are much less trivial than expected, with the ResXNet performance pointing to the fact that it may have significant advantages to its predecessor (ResNet) while dealing with smaller data sets to what concerns generalization and overfitting. 

\subsection{Test-time augmentation}

As previously alluded to, the experiments with the detection threshold were revealing and conducive to better performance on the leaf segmentation but at a cost to the leaf counting task. That may be sufficient to attest state-of-the-art results on the segmentation task, but as the results attested, the performance of rate of positive detection (counting) comes close to being relatively insufficient to be considered a solution. Nevertheless, just because such trade-offs exist, it does not necessarily mean that it can’t be mitigated. 

The solution proposed here is to apply a technique called Test-Time Augmentation (TTA) to mitigate the effects that a different detection threshold causes in the leaf counting task while preserving or even boosting the leaf segmentation performance. In short, test-time augmentation comprises the process of performing inference on the original image and different augmented versions of it. The predictions are then averaged through a pixel-wise voting system and the final output is composed of averaged masks of each instance. As theoretically formulated by \cite{wang2019aleatoric}, the key reason for why it works is that TTA helps to eliminate overconfident incorrect predictions. This is particularly important to the problem of leaf segmentation and counting because it comprises a problem with high occlusion of objects, meaning a combination of a leaf on top of the other is usually classified as a single instance. This is the reason why most results in the challenge leaderboard present a negative difference in the count; they are predicting fewer leaves than are actually present in the images. As an additional note, it is important to highlight that TTA does not influence the process of training in contrast to usual data augmentation; it is only present in the inference part of the model evaluation.

The application of TTA here is composed of definite parts: (i) augmentation of the original image, (ii) inference, (iii) reversing the augmentation on the predictions, (iv) aligning instances of leaves, and (v) averaging the instances masks. The augmentation of leaves is composed of four, simple ones: horizontal and vertical flipping, and 90 degrees rotation clockwise and anticlockwise. This process results in five independent inferences for each image (original plus four augmented versions).  After predictions masks on each of these versions are made, they need to be disaugmented so they can be averaged with the original one. The most challenging part, nevertheless, is that the model is agnostic to the versions of the images so the instance masks in each version are not ordered in the same manner. Hence, an ordering process takes place, which tries to relate the same instance of leaf for all versions of the original image. This is performed by measuring IoU between every instance of the version with the most number of total instances and other versions of the original image. Instances with a value higher than 0.5 are considered the same instance and thus put in the same order for averaging. The fact that the version with the most number of total instances is used in the comparison makes the system to be biased to the version less likely to have committed the error of considering two occluded instances as one. The averaging is somewhat straightforward in the sense that it is a pixel-wise majority voting process; if the mask pixel is positive for a particular instance in more versions that not, the pixel is considered to be existent. Although such a procedure adds processing time to the inference of an image, it is arguably simpler than other methods that need deep technical knowledge to model plants to create synthetic data for training augmentation. The results also point to the fact that it is a reasonable solution for the object occlusion problem leading to expressive results, which might make the performance time trade-off worth it.

\subsection{Programming environment and frameworks}

All the methods described here were performed using accessible and mainly free and open-source frameworks. Regarding the hardware, for example, the Google Colab platform was used. It currently has some time restrictions but offers GPUs of up to 16 GB in memory for free at a limited but considerable amount of time. Colab also has a convenient integration with Google Drive that allows mounting and storage of the data for training. Software-wise, the modelling and training libraries used were part of the Detectron II (\cite{wu2019detectron2}) framework by Facebook, which is a well-maintained, open-source library with pre-trained models available. All the models trained started with pre-trained weights from the COCO data set; such models' weights can be found in the Detectron II's model zoo. The training schedule followed most of the default parameters but with the learning rate set to 0.0001. Checkpoints of the models were saved at every 5,000 interactions and trained until stop converging, which was around 70-100,000 iterations for most. The mainly two augmentation strategies used were horizontal flipping and rotation; for the 101-layer models, the augmentations were random cropping and rotation since flipping the images online consumes much extra memory. The results showed that this did not make a considerable difference while evaluating the models.

\section{Results}
\label{sec:results}

Two points are worth noting before the presentation of the results of the experiments described in the last section. The first concerns results presented regarding the CVPPP data set. They were extracted from the ‘Leaf Segmentation Challenge’ server on CodaLab (\cite{codalab}), and the authors do not have access to the ground truth of this test set. Regarding the Komatsuna data set, it was only used for testing and took no part in the training of the models. The second point relates to how the results and discussions are going to be presented. Since they are from well-defined, independent experiments, the discussions are going to be presented in this section, following each experiment rather than having results and discussions presented in separate sections. 

\subsection{Threshold and certainty}

\begin{figure}[t]
  \centering
  \includegraphics[width=.5\columnwidth]{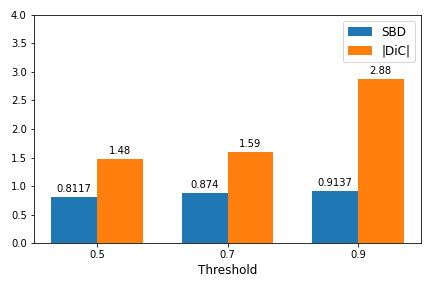}
  \caption{Comparison of the effect of different detection thresholds on the CVPPP test set in the leaf segmentation metric (SBD) and counting (absolute DiC). }
  \label{fig:comp_tresh_coda}
\end{figure}

The effect of the detection threshold is expressive in both the CVPPP and Komatsuna test sets. For the CVPPP, Fig. \ref{fig:comp_tresh_coda} is set to illustrate the comparison between the performance of three key thresholds in the metrics of leaf segmentation (SBD) and counting (abs. DiC). What is most interesting about this comparison is the gains in performance that one makes from a 0.5 to 0.7 threshold. The authors of this paper could not find any discrimination on the previous recent works using deep learning regarding such a threshold, except for \cite{ward2020scalable}, which state to use the standard value of 0.5. The performance gains from going to a 0.7 threshold are so strong that would make a method from ranking low on the leaderboard to one being among the top five. These gains are also strong when compared to the works in the literature that had not published their results as submission in the LSC (\cite{gomes_zheng_2020}). Moreover, going from a threshold of 0.7 to 0.9 would make a method to be in the first place in the leaf segmentation leaderboard but, in this case, with a higher cost to the leaf counting metric. It is interesting to note that the performance gains in the SBD metric are mostly at the cost of a worse performance in leaf counting. That is not necessarily surprising given the previously mentioned points about the trade-off between false and true positive rates, but it is in the sense that such results were achieved by a simple, standard application of Mask R-CNN only using flip and rotation as augmentation strategies. However, one can also notice from Fig. \ref{fig:comp_tresh_coda} that the gains are not linear; a strong gain in the SBD metric from 0.5 to 0.7 does not cost as much in the abs. DiC metric as the move from 0.7 to 0.9. 

\begin{figure}[t]
  \centering
  \includegraphics[width=.55\columnwidth]{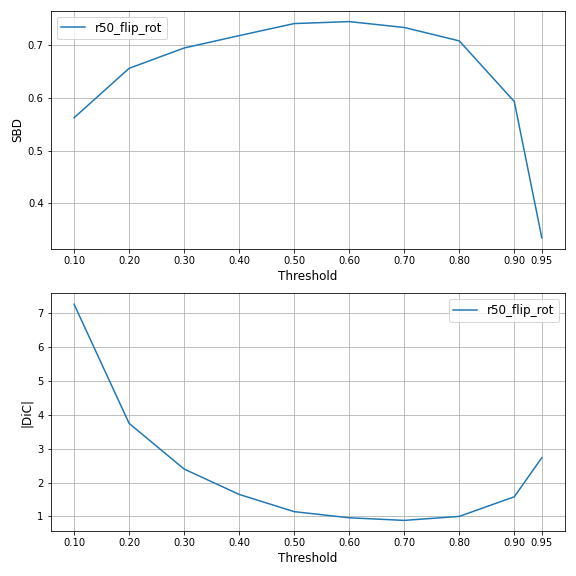}
  \caption{The effect of different thresholds on the Komatsuna test set in the SBD and abs. DiC metrics.}
  \label{fig:comp_tresh_komat}
\end{figure}

While testing different thresholds on the Komatsuna data set, however, results showed that the performance gains of the different thresholds might be somewhat overly expressed in the CVPPP data set. Fig. \ref{fig:comp_tresh_komat} is set to illustrate the difference in the SBD and DiC in detail. The results show that although the detection threshold can have a great influence on such an external data set, the leaf segmentation metric does not monotonically increase to from 0.5 to 0.9, as seen in the CVPPP. This fact is remarkably insightful, pointing to the fact that the CVPPP might represent a narrow data distribution that can greatly affect model generalisation. In simpler terms, the SBD monotonically increases due to the fact that the examples in the test set are too similar to the ones in the training set. It should thereby raise some concerns to what it means to optimize an object segmentation metric on the CVPPP data set, which has become somewhat of a benchmark in this field, in relation to other data sets for the same task. The best SBD achieved in the Komatsuna data set was 0.7452, with the model obtaining an abs. DiC of 0.9594 at a threshold of 0.6. This result is also noteworthy since it surpasses the one achieved by the previous work with the highest SBD score, at the best of our knowledge, which was 0.7169 (\cite{ward2020scalable}), when no images from the data set (out of context) are used.

\subsection{Model depth and cardinality}

\begin{figure}[t]
  \centering
  \begin{subfigure}[t]{0.5\columnwidth}
    \includegraphics[width=\textwidth]{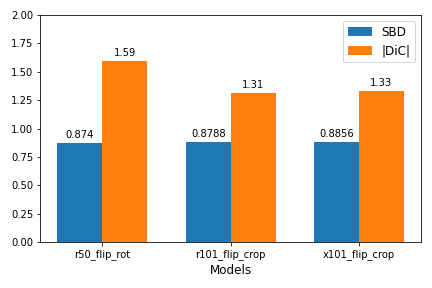}
    \caption{}
  \end{subfigure}             
  
  \begin{subfigure}[t]{0.5\columnwidth}
    \includegraphics[width=\textwidth]{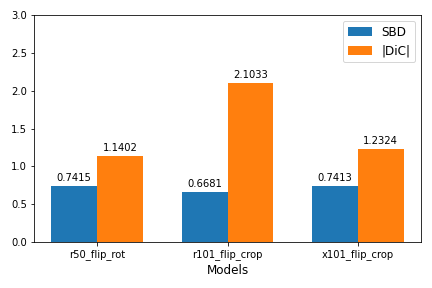}
    \caption{}
  \end{subfigure}
  \caption{Comparison of performance of models of different depth and cardinality on the tests sets: a) CVPPP and b) Komatsuna. Detection threshold = 0.7.}
  \label{fig:comp_dep_card}
\end{figure}

A consequent question is how models with different depth will perform in these tasks while trained by the same algorithm. The results from such an experiment were also revealing; Fig. \ref{fig:comp_dep_card} is set to illustrate some comparisons. In this case, the figure compares the 50- and the two 101-layer models (ResNet and ResXNet), all for the same threshold of 0.7. The first image (a) illustrate that there are small differences from the models of 50 and 101 layers with the ResNet backbone regarding the segmentation but a performance gain in the leaf counting task. In advantage, the ResXNet backbone seems to have significant gains in both segmentation and count. However, the Komatsuna data set can be used again to clarify the validity of the potential gains. In the second image (b), it is clear that the ResNet backbone with 101 layers actually losses a lost of performance in both tasks, while little difference is seen on the ResXNet. This comparison reveals important insights regarding this characteristic called cardinality, which in this case, stopped the deeper model to overfit as it happened with the ResNet of the same depth.  It points to the fact that such characteristic — the only difference between these backbones — may work better in problems of objects with high occlusion, allowing the model to generalise significantly better to other data sets. Lastly also pointing to the mentioned fact that the data distribution from the CVPPP might be too narrow to result in strong generalisation in other data sets, as any model can easily overfit to it and perform worse when tested on external data sets.

\subsection{Test-time augmentation}

\begin{figure}[t]
  \centering
  
   \centering
  \begin{subfigure}[t]{0.55\columnwidth}
    \includegraphics[width=\textwidth]{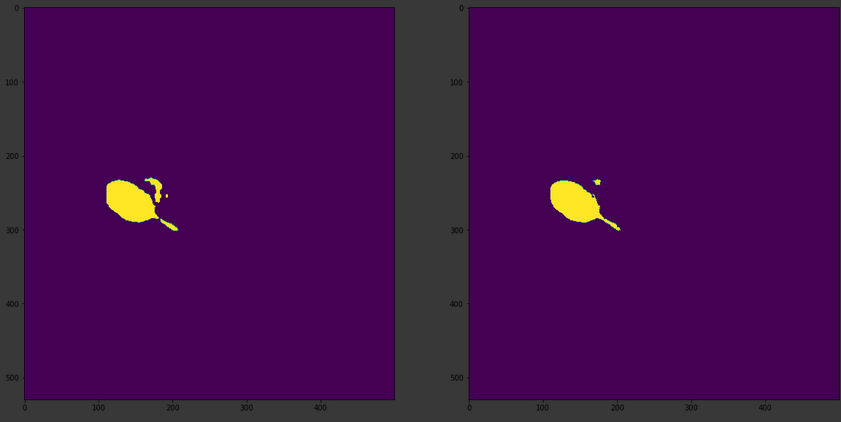}
    \caption{Single instance mask.}
  \end{subfigure}  
  
  \centering
  \begin{subfigure}[t]{0.55\columnwidth}
    \includegraphics[width=\textwidth]{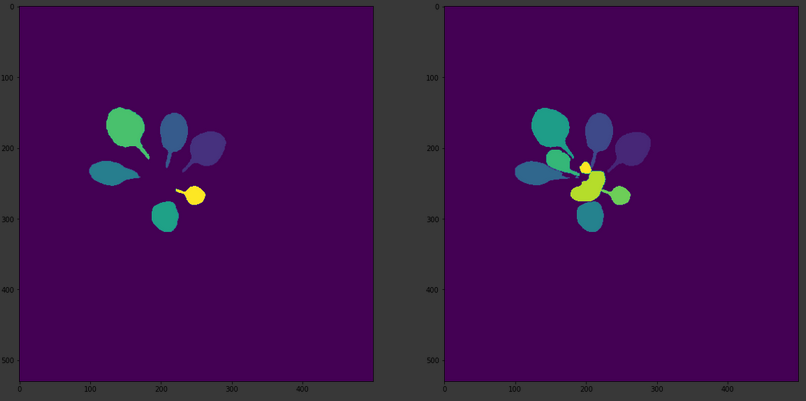}
    \caption{All instances.}
  \end{subfigure}             
  
  \begin{subfigure}[t]{0.55\columnwidth}
    \includegraphics[width=\textwidth]{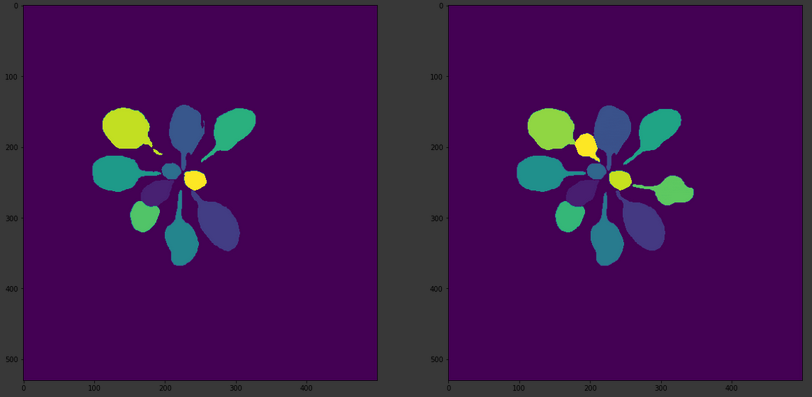}
    \caption{All instances.}
  \end{subfigure}
 \caption{Examples of instances where there was an expressive difference from not using (left) and using (right) TTA.}
  \label{fig:tta_app}
\end{figure}

\begin{figure}[t]
  \centering
  \begin{subfigure}[t]{0.5\columnwidth}
    \includegraphics[width=\textwidth]{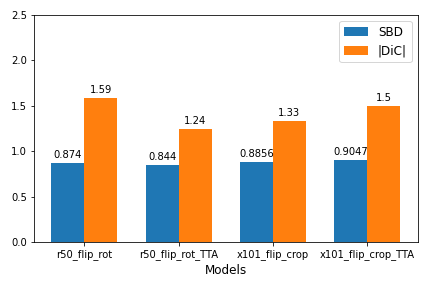}
    \caption{}
  \end{subfigure}             
  
  \begin{subfigure}[t]{0.5\columnwidth}
    \includegraphics[width=\textwidth]{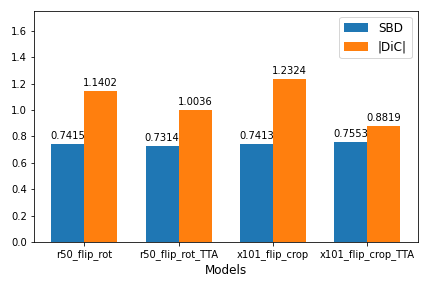}
    \caption{}
  \end{subfigure}
  \caption{TTA performance comparison in the 50-layer ResNet backbone and the 101-layer ResXNet backbone on a) the CVPPP and b) Komatsuna data set.}
  \label{fig:comp_TTA}
\end{figure}

The test-time augmentation hypothesis is proposed here as a way to alleviate the trade-off or maximize the relationship between the segmentation and counting scores. A classifier that is able to segment leaves but only does it for the ones that it has a high certainty of detection is not as useful, despite higher SBD score. To that end, the experiment showed to be consistent for both data sets but it showed to be more useful for ResXNet backbones as illustrated in Fig. \ref{fig:comp_TTA}. The model 101-layer ResXNet model, which is currently submitted in the LSC, have achieved approximately 2\% higher performance in the SBD metric than other candidates while having a more reasonable counting performance than other thresholds that achieve comparable performance but do not use TTA. To illustrate the effect that such a technique can have on segmenting the masks, Fig. \ref{fig:tta_app} shows examples of segmentation instances where the effects of using TTA are expressive. 

This result becomes more important when the Komatsuna is considered, which most methods in the literature do not take into account. As shown in Fig. \ref{fig:comp_tresh_coda}, this performance in the SBD metric can be achieved by forcefully increasing the threshold of detection. However, as shown in Fig. \ref{fig:comp_tresh_komat}, this result does not translate to the external data set (Komatsuna), which presents very poor performance for a higher threshold with no TTA. Therefore, by applying TTA, the model not only makes a better compromise between segmentation and counting on the CVPPP data set but also generalises much better in the Komatsuna data set. It achieves the highest result observed in both segmentation and counting metrics on this external data set with an SBD of 0.7553 and an abs. DiC of 0.881 (higher than the CVPPPP — the test set closely related to the training set).

\subsection{Comparison with previous and LSC methods}

\begin{figure}[t]
  \centering
  \begin{subfigure}[t]{0.4\columnwidth}
    \includegraphics[width=\textwidth]{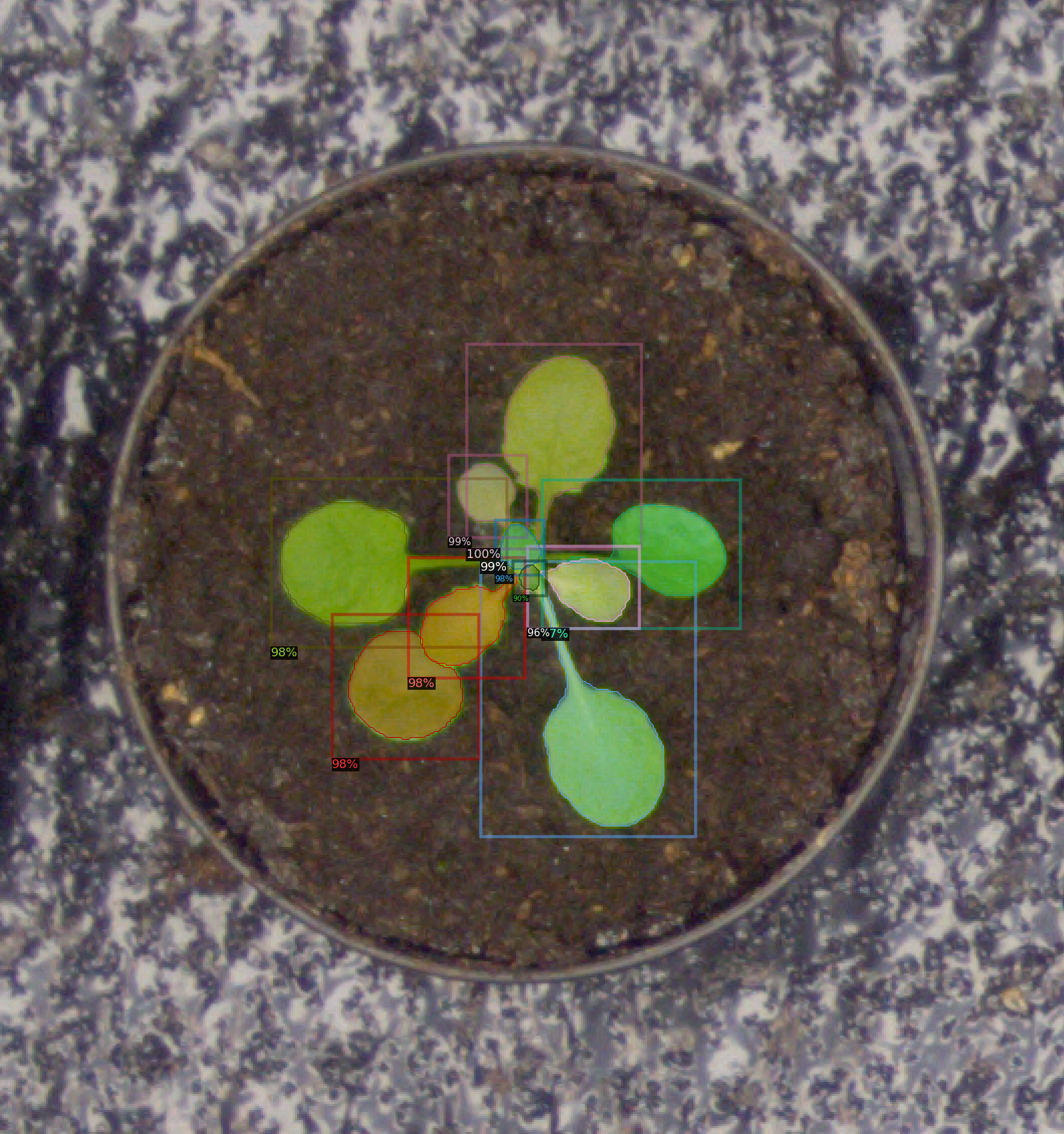}
    \caption{}
  \end{subfigure}             
  \begin{subfigure}[t]{0.4\columnwidth}
    \includegraphics[width=\textwidth]{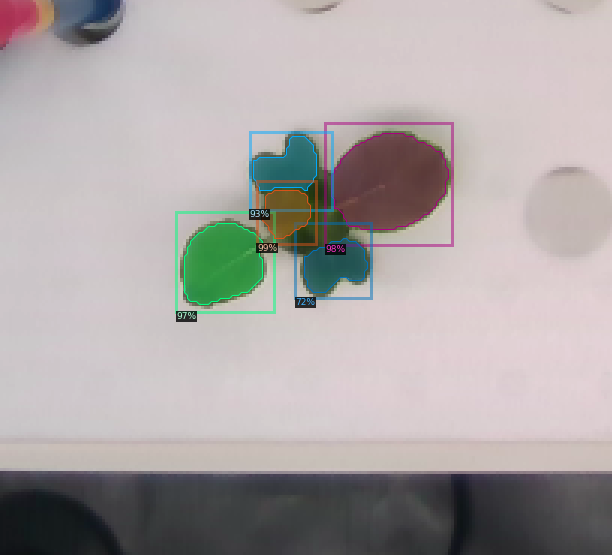}
    \caption{}
  \end{subfigure}
  \caption{Examples of inference on the images from the (a) CVPPP and (b) Komatsuna data sets.}
  \label{fig:inf_ex}
\end{figure}

The analyses and results presented attest that there are different ways to present a model, especially when considering the trade-off between the tasks of segmentation and counting. Nevertheless, given the strategies used here for mitigating the costs of such trade-off, the resulting models present competitive results for both tasks despite the bias adopted. Table \ref{tab:sbd_comp}, for example, shows the leaderboard of the LSC when sorted by the SBD metric. When biased towards segmentation (higher threshold), the model performs 2\% higher than the second place while demonstrating competitive scores for leaf counting. If the threshold is biased towards the task of counting, however, the model can tie the results with the best-performing model in the abs. DiC metric while also presenting the same score for SBD as the best submission for the SBD metric. As illustrated in Table \ref{tab:dic_comp}, the best submission for leaf counting have lower performance at segmenting the leaves than the one proposed here. It should be noted that there is an advantage in tying when the model proposed here does not use any synthetic and additional data in training and uses a straightforward Mask R-CNN algorithm. The comparison of the Komatsuna data set with another method is presented in Table \ref{tab:komat_comp}. It is expected that, as this is a relatively recent data set, more works will present their results using this data set as a benchmark. It is also worth remembering that the Komatsuna data set was not used at any point while training the model but only while testing it. Finally, examples of inferences in the CVPPP and Komatsuna data sets are illustrated in Fig.  \ref{fig:inf_ex}.

\begin{table}
	\caption{Top 5 LSC leaderboard sorted by SBD score (segmentation biased)}
	\centering
	\begin{tabular}{lll}
		\toprule
		Name (user)     &  SBD      &  |DiC|  \\
		\midrule
		\bfseries{dougpsg (ours, segment. biased)} & \bfseries{0.90} & \bfseries{1.50}      \\
		LfB & 0.88  & 3.97     \\
		wuzi     & 0.87       & 0.9  \\
		yukinohana     & 0.87       & 0.89  \\
		looooong     & 0.86       & 1.24  \\
		\bottomrule
	\end{tabular}
	\label{tab:sbd_comp}
\end{table}

\begin{table}
	\caption{Top 5 LSC leaderboard sorted by |DiC| score (count biased)}
	\centering
	\begin{tabular}{lll}
		\toprule
		Name (user)     & |DiC|     &  SBD  \\
		\midrule
		yukinohana & 0.89  & 0.87     \\
		\bfseries{dougpsg (ours, count biased)} & \bfseries{0.89} & \bfseries{0.88}      \\
		wuzi     & 0.9       & 0.87  \\
		looooong     & 1.24       & 0.86  \\
		StevenYuan     & 2.74       & 0.76  \\
		\bottomrule
	\end{tabular}
	\label{tab:dic_comp}
\end{table}

\begin{table}
	\caption{Komatsuna comparison}
	\centering
	\begin{tabular}{lll}
		\toprule
		Work      &  SBD      &  |DiC|  \\
		\midrule
		\bfseries{Ours, segment biased} & \bfseries{0.75} & \bfseries{0.88}      \\
		\cite{ward2020scalable}    & 0.72       &  \\
		\bottomrule
	\end{tabular}
	\label{tab:komat_comp}
\end{table}

\section{Conclusions}

This paper presents a series of experiments, which culminates on a model achieving state-of-the-art results for the task of leaf segmentation while presenting competitive results for leaf counting. At the same time, differently from most recent methods in the related literature, such results are achieved without complicated pipelines for synthetic data generation that attempt to augment the limited CVPPP training set. While built and trained on an accessible platform without significant modifications to the standard Mask R-CNN framework, this model represents a simpler solution that achieves best results in both the LSC and on the external data set (Komatsuna). Most insights came from addressing under-discussed topics such as the effect of the detection threshold on the task at hand and ways to mitigate the trade-off resulted from it. The best results came from a deeper model (101 layers) with different cardinality (ResXNet), which could use the benefits of performance gains and trade-off mitigation given by test-time augmentation. One can argue that these aspects are significantly relevant to tasks like leaf segmentation, which are composed of objects of single class that suffer from high occlusion. The experiments also generated additional insights about the data set, confirming potential concerns that the segmentation and counting metrics should not be considered alone, as they can be forced by changes in the threshold, and that using data sets of narrow distributions to benchmark solutions may lead to erroneous conclusions regarding the generalisation of the resulted models.

\bibliographystyle{unsrtnat}
\bibliography{references}  






\end{document}